\documentclass[pdflatex,sn-mathphys-num]{sn-jnl}

\usepackage{graphicx}%
\usepackage{multirow}%
\usepackage{amsmath,amssymb,amsfonts}%
\usepackage{amsthm}%
\usepackage{mathrsfs}%
\usepackage[title]{appendix}%
\usepackage{xcolor}%
\usepackage{textcomp}%
\usepackage{manyfoot}%
\usepackage{booktabs}%
\usepackage{algorithm}%
\usepackage{algorithmicx}%
\usepackage{algpseudocode}%
\usepackage{listings}%
\usepackage{natbib}%
\theoremstyle{thmstyleone}%

\theoremstyle{thmstyletwo}%

\theoremstyle{thmstylethree}%

\begin{document}

\title[Article Title]{\textbf{Engineering Sentience}}

\author*[1]{\fnm{Konstantin} \sur{Demin}}\email{deminkasci@gmail.com}

\author[2]{\fnm{Taylor} \sur{Webb}}\email{taylor.w.webb@gmail.com}

\author[3,4]{\fnm{Eric} \sur{Elmoznino}}\email{eric.elmoznino@mila.quebec}

\author*[1,5,6]{\fnm{Hakwan} \sur{Lau}}\email{hakwan@gmail.com}

\affil[1]{\orgdiv{Center for Neuroscience Imaging Research}, \orgname{Institute for Basic Science}, \orgaddress{\city{Suwon}, \country{South Korea}}}

\affil[2]{\orgname{Microsoft Research}, \orgaddress{\city{New York}, \country{USA}}}

\affil[3]{\orgname{Mila - Quebec AI Institute}, \orgaddress{\city{Montréal}, \country{Canada}}}

\affil[4]{\orgname{Université de Montréal}, \orgaddress{\city{Montréal}, \country{Canada}}}

\affil[5]{\orgdiv{Department of Biomedical Engineering}, \orgname{Sungkyunkwan University}, \orgaddress{\city{Suwon}, \country{South Korea}}}

\affil[6]{\orgdiv{Department of Intelligent Precision Healthcare Convergence}, \orgname{Sungkyunkwan University}, \orgaddress{\city{Suwon}, \country{South Korea}}}

\abstract{We spell out a definition of sentience that may be useful for designing and building it in machines. We propose that for sentience to be meaningful for AI, it must be fleshed out in functional, computational terms, in enough detail to allow for implementation. Yet, this notion of sentience must also reflect something essentially ‘subjective’, beyond just having the general capacity to encode perceptual content. For this specific functional notion of sentience to occur, we propose that certain sensory signals need to be both assertoric (persistent) and qualitative. To illustrate the definition in more concrete terms, we sketch out some ways for potential implementation, given current technology. Understanding what it takes for artificial agents to be functionally sentient can also help us avoid creating them inadvertently, or at least, realize that we have created them in a timely manner.}

\keywords{subjective processing in AI, functional sentience, perception-reality monitoring, qualitative perception}

\maketitle

\section{Introduction}\label{sec1}

Recent advances in artificial intelligence (AI) research have sparked renewed controversy as to whether machines can be sentient. One commonly acknowledged problem is that we lack a broad consensus on how to define the term ‘sentience’. Our goal here is to develop a workable approach to the concept of ‘sentience’ - which we call functional sentience - for AI research and to discuss its possible implementations. This approach seeks to bridge the gap between philosophical debates and practical AI system design, grounding the concept in computational frameworks that are directly applicable to AI development.

An apparent dilemma is that authors are often either defining sentience in metaphysical terms (using non-empirical concepts that go beyond normal science) \cite{Roelofs2021-qm, Fleming2023-nf} or are defining it in terms of relatively trivial functional processes, e.g. by stipulating that sentience or consciousness is just to make perceptual information globally available within the system \cite{Dehaene2011-jr}. The former is beyond the scope of our present discussion. For the latter, the relevant mechanisms are easy to implement, e.g. multimodal AI systems will satisfy the requirements of some of them \cite{Butlin2023-jf}. However, we think they do not capture the essentials of subjective human perceptual experiences.

To illustrate this problem, consider that the global workspace theory says that a piece of (sensory) information becomes conscious if it is broadcasted by some central mechanism to be shared with submodules within the entire network \cite{Baars1993-lk}. However, such a mechanism is useful for a wide variety of general cognitive functions besides sentience, such as thinking, remembering, talking, etc. As such, the reduced neuronal activity (associated with the broadcast) in the global workspace in unconscious states (e.g. coma) may just reflect a subject's state of general cognitive impairment rather than a lack of subjective experiences \textit{per se}.

On the other hand, people with aphantasia also don’t report the experience of vivid mental imagery, but nevertheless can perform tasks that are associated with global workspace functions (e.g. mental rotation of objects) \cite{Kay2024-rb, Zhao2022-fe, Pounder2022-uc, Michel2025-cj}. Similar problems apply to many other theories of consciousness that hinge on some type of processing thought to be \textit{generally} sophisticated and advantageous. This lumps up together sentience or consciousness with reduced processing capacity in general.

Here we take a different approach. We define \textbf{\textit{functional sentience}} to mean that an agent’s perceptual experiences serve a \textit{specific} function for that agent, rather than just to enable the agent to complete simple and common perceptual tasks in general. In other words, to be functionally sentient in this sense is to be able to perform a \textit{specific} computation that serves a specific, predefined purpose. An agent that lacks sentience may be able to perform some perceptual tasks, but should fail specifically for tasks where nonconscious processing in humans seems insufficient \cite{Weiskrantz1997-kx, Aleci2024-mh, Ajina2016-zk, Overgaard2012-xb}. For that, we stipulate that all relevant aspects of sentience can be explained in specific terms of computation and the implementation of software algorithms. In philosophical terms, this is to say that we assume functionalism \cite{Polger2012-sl} for this purpose. We do not claim that our definition is the only one available, but rather we strive to provide an example of how one can approach the question of sentience in a way that is coherent and meaningful for studying perceptual experience in humans, AI systems, and other non-human agents.

\section{Functional Origins: Perception-Reality Distinction}\label{sec2}

To understand the computational function of sentience for an AI agent, it may be useful to consider its possible biological and evolutionary origins. Here, we hypothesize that functional sentience might have evolved in some animals for the sake of maintaining the distinction between appearance and a model of objective reality. We also argue that this biological insight provides a foundation for designing AI models that can similarly distinguish between sensory inputs and a constructed representation of the external world.

The actions of some simple creatures are directly linked to their senses. The ‘rules’ governing their responses to stimuli are relatively simple and inflexible, like reflexes. These sensory signals can be described as \textbf{assertoric}, in the sense that they are persistent and unignorable. Simple creatures like insects behave as if they take their sensory signals to be infallible reflections of the world, in such a way that they are incapable of doubting the validity of these signals, regardless of data from other sensory systems or their knowledge. However, many animals, including adult humans, seem to establish knowledge regarding the state of the world that can dissociate their beliefs from the sensory data. We can call this (potentially non-veridical) representation of reality a Reality Model. For AI, constructing such a Reality Model involves creating a dynamic internal representation that can update and reconcile discrepancies between raw sensory data and established knowledge. To very simply formalize it, $Z_t$ is an agent's Reality Model that represents the environment's state at time $t$, which could be a vector, matrix, or other structured data type that captures relevant aspects of the environment.

How should an agent treat a perceptual signal when its content conflicts with the pre-existing Reality Model? One simple mechanism may be to weigh the different sources of information by their estimated reliability. If the perceptual signal is (judged to be) strong and reliable relative to the certainty of the model, one may update the model with the content of the perceptual signal. In contrast, if the perceptual signal is (judged to be) weak and unreliable, one may continue to maintain the content of the model and ignore the perceptual signal. In most cases, the certainty for the two may be more balanced, so that one would combine the two sources of information depending on the relative reliability of both, and come to a weighted average as a conclusion, in order to update the model accordingly. In AI, this could translate to algorithms that integrate sensor data with Bayesian reasoning models, allowing the system to continually refine its understanding of the environment by updating its knowledge in light of new evidence \cite{Papamarkou2024-ig}.

In other words, sentience might have evolved as a way for the agent to maintain what psychologists call the perception-reality distinction \cite{Dijkstra2022-ty}. For AI, achieving a perception-reality distinction could mean developing algorithms capable of resolving conflicts between sensory inputs and predictive models, allowing for adaptive behavior in complex environments. There may be advantages to never completely ignoring or writing off the perceptual signals, however wildly they contradict the content of the Reality Model.

For example, if an agent is red-green colorblind, it may be useful to know that red and brown things may be easily confused with one another \textit{to oneself}, whereas in reality, they may be very different and others may not see them as similar at all. Such awareness can help an agent anticipate errors likely to be committed by oneself while maintaining correct knowledge about the world and other agents. Similarly, an AI system might be designed to recognize its own sensory limitations and adjust its behavior accordingly, enhancing its reliability in varied contexts.

In contrast to other proposed functions of sentience, the current proposal represents a relatively specific notion that applies only to some creatures and agents with certain overall cognitive architectures (such as having something akin to a Reality Model, coupled with ongoing perceptual processes). This helps us set the stage for considering in some detail how functional sentience could be defined and implemented with technology currently available or conceivable soon. The below is not meant to be a definitive implementation plan, but rather a sketch to illustrate that \textit{some} implementation should be feasible. As proposed in the introduction, the goal here is not to develop a detailed recipe but rather to illustrate why these concepts are potentially useful for understanding sentience in agents beyond humans.

\section{Defining Functional Sentience}\label{sec3}

\begin{figure}[h]
\centering
\includegraphics[width=0.9\textwidth]{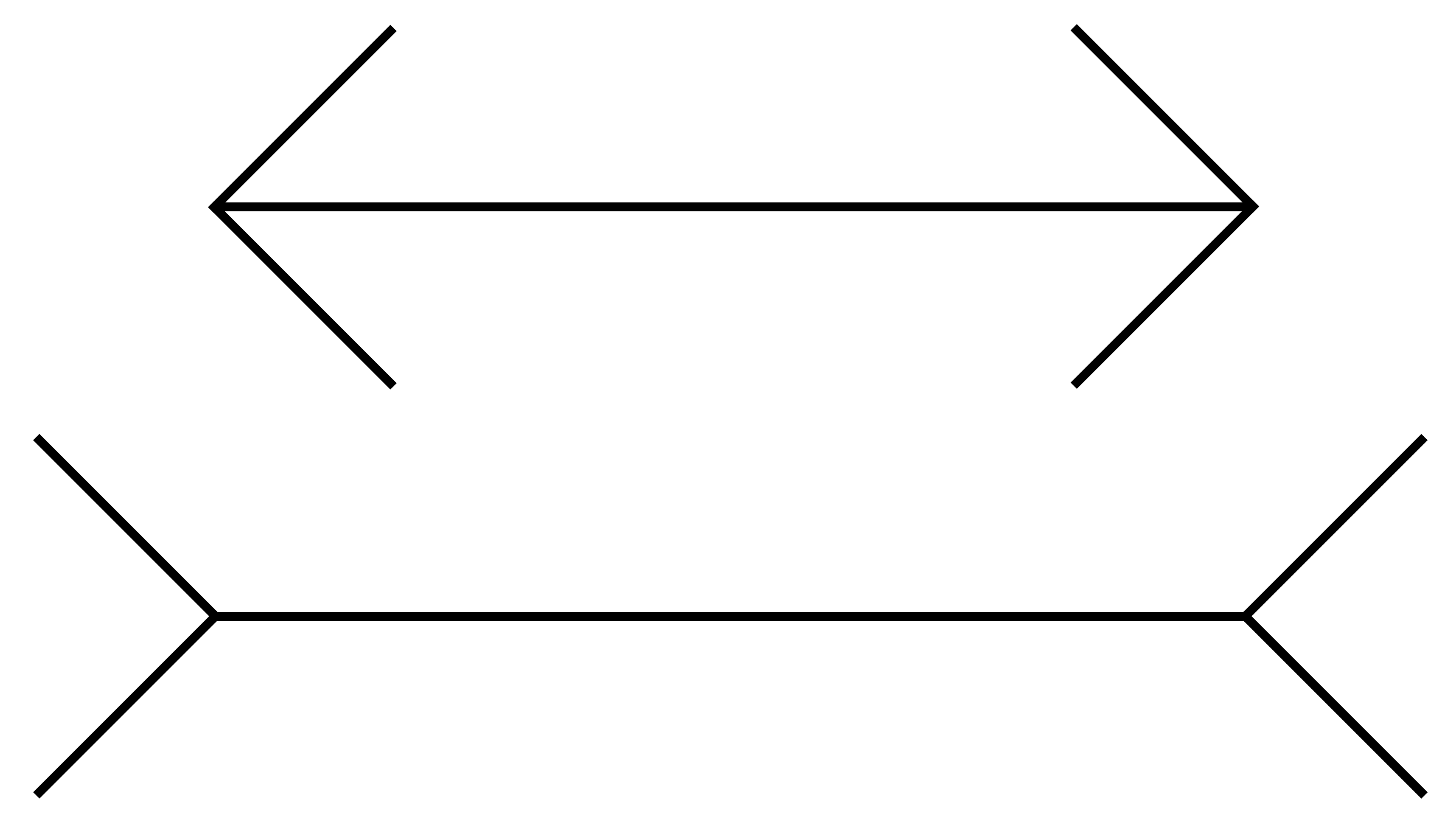}
\caption{Müller-Lyer illusion \cite{Muller-Lyer1889-qt}. The line segment that has arrows pointing inwards appears longer than the line segment that has arrows pointing outwards, but in reality, those line segments are the same length. Note that the illusion doesn’t disappear even if you \textit{know} that the arrows are of the same physical length, i.e., it is \textit{assertoric}.}\label{fig1}
\end{figure}

We now define functional sentience as the capacity for the meaningful processing of sensory signals that are both assertoric and qualitative. Here we will explain in detail what we mean by that and in the upcoming sections we will sketch out how this could be meaningfully implemented in some AI systems to mimic key aspects of sentient experience.
As mentioned above, by having assertoric signals, we mean that the system treats the relevant sensory signals as \textit{prima facie} reflecting the current state of the world, in a way that is not easily ignorable. Sentient agents, like some other simpler agents, have automatic sensory inputs that strongly affect their subjective knowledge but do not directly alter them in a deterministic manner. Those feelings persist even in the face of contradictory evidence from other sources. For example, the feeling of sharp pain in one’s finger cannot be reasoned away, even if one \textit{knows} that there is, in fact, nothing wrong with one’s finger. A similar effect can be observed in visual illusions (Fig. \ref{fig1}). These feelings can’t be turned off, meaning that such an agent lacks a pronounced top-down effect on the sensory inputs. Formally, in its simplest form, we can define an assertoric signal $S$ as one where the processing function $P(S)$ assigns a high priority or weighting to $S$ in the system's decision-making process, irrespective of conflicting data $C$ from other sources:

\begin{equation}
P(S)=max(\alpha S+\sum^n_{i=1}(\beta_iC_i), 1), \nonumber
\end{equation}

where $\alpha$ represents the priority coefficient for the sensory signal $S$, and $\beta_i$ represents the weighting of each conflicting input $C_i$. For assertoric signals, $\alpha$ is set significantly higher than $\beta_i$, ensuring that $S$ dominates the decision process.

In practice, AI systems could implement this by using a neural network where the output layer assigns a higher weight to certain types of sensory inputs, effectively making them non-ignorable in the network's output. From this standpoint, we may also assume that an agent must obtain strong knowledge to affect behavior significantly. Alternatively, we can also completely disregard any effect of conflicting inputs at the point of signal acquisition and suggest that the difference is in the fact that conflicting inputs can affect the behavior only in the late stages of signal processing.

\begin{figure}[h]
\centering
\includegraphics[width=0.7\textwidth]{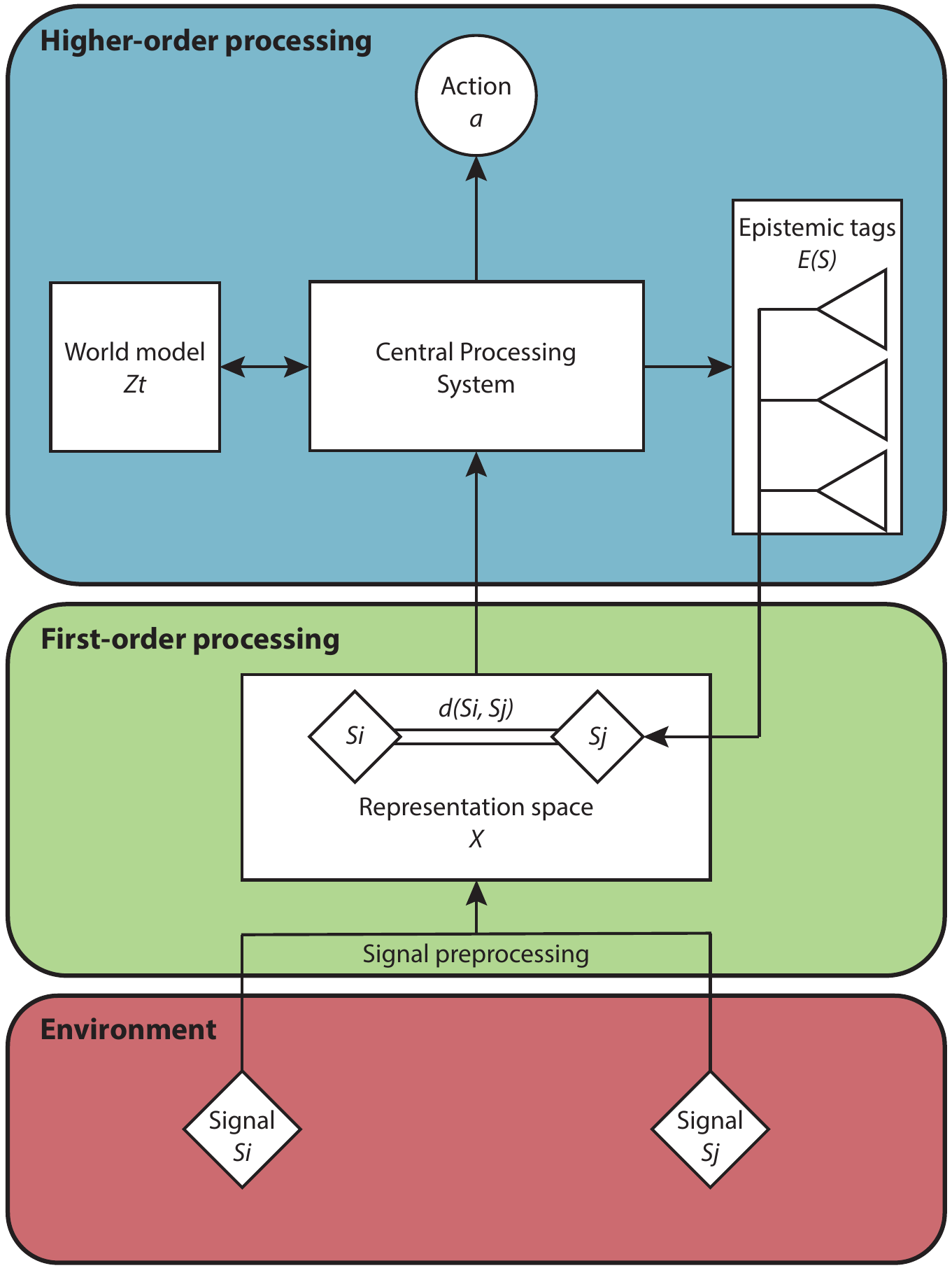}
\caption{A flow chart summarizing the overall architecture for the implementation of functional sentience.}\label{fig2}
\end{figure}

By ‘qualitative’, we mean sensory signals that are in a format such that they are i) \textit{immediately available} for an agent to reflect and act upon and that they ii) \textit{reflect the agent’s own state} (i.e. not someone else’s state) and iii) are encoded in a well-defined \textit{similarity structure} \cite{Beck2019-jn, Clark2000-ki} (Fig. \ref{fig2}). To put the last point (iii), in other words, qualitative signals are non-categorical and high-dimensional. The similarity structure requirement allows for the representation of a perceptual experience in terms of its degree of pairwise similarity to all other available experiences of this modality. For example, red is not just different from blue, it is also more similar to pink than to (some shade of) purple to a certain degree, etc. We call such representations the Similarity Profile. The representation need not be continuous; it could be discrete, like a graph or a finite set of vectors, as long as there is a clear and consistent way to measure similarity between different states.

To define it, let $S_i$ and $S_j$ represent two sensory signals within a structured space $X$ that possesses a well-defined similarity structure. The similarity between these signals can be captured by a distance function $d(S_i, S_j)$, which might be Euclidean in the case of continuous representations, or another metric appropriate for the structure at hand. The key requirement is that this function provides a way to compare the similarity of different sensory inputs, such that $d(S_i, S_j)$ is small for similar signals and large for dissimilar ones.

These Similarity Profiles must be immediately available to something we call the Central Processing System (defined below), meaning they are represented in a certain format, such that access to them is automatic - this refers to point (i) above. When we see red, we see it more similar to pink than to blue without any additional effort required. It just looks like that. This requirement implies that the Central Processing System can directly operate on the encoded signals within their structured representation without needing to access raw sensory data or external processing steps. In short, the Central Processing System refers to the core computational mechanism within an AI agent that processes sensory inputs, applies logical reasoning, and generates decisions or actions. It acts as the central hub where all relevant information is synthesized to form coherent, goal-directed behavior. Central Processing System output is typically a decision or action, which could be represented by a choice of action $a$ from a set of possible actions $A$, based on maximizing a utility function $U$:

\begin{equation}
a*=argmax_{a\in{A}}U(a\:|\:inputs\:from\:CPS)
\nonumber
\end{equation}

Finally, similarity profiles reflect only the nature of one’s own sensory signals themselves, rather than general knowledge of the world (point ii above). For example, some current large language models are capable of making ‘correct’ statements such as “red is more similar to pink than to blue” \cite{Marjieh2023-jz, Kawakita2024-cp}. Congenitally blind individuals might also have abstract knowledge of this kind. But to the extent that these sorts of judgements reflect actual ‘knowledge’, this is information about the stimuli and how they are perceived by others, rather than information about how those stimuli are represented within the system itself. Thus, in the context of functional sentience, it’s crucial that these profiles are grounded in the system’s internal sensory data and structured in a way that reflects the system’s own experiences, not just abstract knowledge.

In sum, functional sentience concerns the processing of \textit{assertoric} and \textit{qualitative} sensory signals. We illustrate below why this definition may be useful and suitable for making progress in AI research, and at the same time captures the central aspects of sentience as we know it. 

\section{Implementation of assertoric Sensory Signals}\label{sec5}

As suggested above, sentience may have evolved out of the need to maintain sensory information in parallel with knowledge about the world. We further argue that sentience can be thought of as a mechanism with which perceptual information communicates with the Central Processing System. Without such a system, there would have been no point in implementing the sentience that we discussed here. Until recently, the implementation of the Central Processing System in machines may seem like a very demanding challenge. However, the advancements in research on large language models (LLMs) might have changed that outlook quite considerably. While many believe that we are still far from artificial general intelligence \cite{Feng2024-vz, Morris2023-xf} and that current LLMs evidently still make many trivial mistakes in terms of reasoning \cite{Prabhakar2024-eq, Alamia2023-bw}, these obstacles have been made considerably more tractable \cite{Wei2022-cx, Chia2023-pa}.

One key element in our definition is that the relevant sensory signals need to be presented to the Central Processing System as \textit{assertoric}. Assertoric sensory signals can be implemented simply by associating a reliable signal with a categorical (symbolic) representation in the Central Processing System, indicating that the signal correctly reflects the state of the world. We can call these categorical representations Epistemic Tags, as one of their functions is to specify the state of informativeness of specific sensory signals. After processing the sensory signals and previous knowledge as premises in a logical argument, the conclusion about signal reality should be rationally drawn by the processing system, and the corresponding Epistemic Tag should be attached. The Epistemic Tag function may be necessary for agents with imperfect perceptual capacities because sometimes the system would fail, and it should commit to the signal only if some computational process judges it to be reliable above a certain threshold. 

As an example of implementation, let $S$ represent a sensory signal and $E(S)$ represent its associated Epistemic Tag, which encodes the reliability or informativeness of the signal. The Central Processing System processes sensory inputs by combining $S$ and $E(S)$ into a decision-making function $D$:

\begin{equation}
D(S,E(S))=argmax_i(f_i(S)E(S)),
\nonumber
\end{equation}

where $f_i(S)$ represents different potential interpretations or inferences that can be drawn from the signal $AS$, and the Epistemic Tag $E(S)$ serves as a weighting factor that influences which interpretation is selected. The choice of the exact function depends on how we want our epistemic tags to influence the final model decision and could be potentially adjusted. As an example of how this model works, let’s come back to the Müller-Lyer illusion discussed above (Fig. \ref{fig1}). In this case, the signal interpretation of the lines being the same size fails to exceed the interpretation of them having different sizes, despite the attached epistemic tag that informs the system of the mistake.

Furthermore, these Epistemic Tags may also serve some other functions, allowing for any kind of variable binding with the appropriate sensory representations. To illustrate, let us consider an alternative architecture in which the sensory signals are categorized into categorical representations that carry an object recognition result to be output to the Central Processing System. In that case, the sensory representation of a yellow cat may be symbolically represented as a [Yellow Cat]. Once information reaches this level, the Central Processing System can make inferences without referring back to the sensory signals. 

This way, the Central Processing System can indirectly access the correct source of information without duplicating fine-grained information, like a proxy object that acts as a pointer to real sensory data and has epistemic metadata attached to it, that manages its access but doesn’t interact with data directly. In formal terms, the Central Processing System operates on a higher-level abstraction $\phi(S)$, which is a function that maps the raw sensory signal $S$ to its categorical representation. This allows the system to perform reasoning tasks using these categorical forms, which are enriched with epistemic metadata but do not require direct interaction with the raw data.

Here, a vast number of known deep learning solutions can be implemented to reproduce these assertoric signals in machines. Importantly, all of these solutions share a simple architectural design, closely resembling the functional sentience architecture suggested earlier (Fig. \ref{fig2}). As is standard in deep learning, a (first-order) neural network takes sensory data and/or top-down signals as input and produces a number of perceptual representations distributed across a hierarchy of layers. Thus, outputting $f_i(S)$ value of inferences that can be drawn from the signal $S$.

In parallel, a series of separate (higher-order) neural networks can each take a first-order layer’s activations as input and then output a single scalar, representing the probability that the first-order representation of that layer is veridical considering the Reality Model (or assigning any other properties that we will omit from further discussion here). These scalar probabilities can, in effect, be used as epistemic tags $E(S)$ by the rest of the system. Solutions then differ primarily in terms of how the second-order networks are trained and how the epistemic tags that they output are used.

If supervision signals are occasionally present that would provide the second-order networks with a ‘ground-truth’ about the reliability of their first-order representations, then the second-order network can be trained to estimate the probability of correctness by standard supervised learning. For example, if a source of first-order representation errors is internal noise in the network, ground truth can be estimated simply by averaging noisy first-order representations over time. Another possibility is to obtain ground truth by comparing representations of the same percept but across different sensory modalities (e.g., verifying the veracity of a sound using visual feedback) or through movement (e.g., checking if a visual percept behaves as it should, given known motor actions).

If ground truth is not directly available, the second-order networks can be trained on other surrogate tasks where the reliability of a signal is an implicit factor in performance. For instance, externally generated signals can often be more predictable than certain kinds of false, internally generated signals (e.g., hallucinations arising from random internal noise). The second-order networks might then try to predict upcoming first-order representations using past ones. If the prediction has high error (i.e., high surprisal), then the second-order network can assign a lower probability to the veracity of the first-order representation. 

In practice, this approach can be summarized as follows:

\begin{equation}
Prediction\:error=\lVert{h(S_t)-\hat{h}(S_t)}\lVert^2,
\nonumber
\end{equation}

where $h(S_t)$ is the actual first-order representation of a signal $S$ at time $t$ and $\hat{h}(S_t)$ is the predicted representation based on past inputs. The higher the prediction error, the lower the epistemic reliability assigned to the signal. This approach is similar to the idea of predictive coding and active inference, which has already seen applications in modern deep learning (e.g., \cite{Millidge2022-to, van-den-Oord2018-wg, Alamia2023-bw}), except that here the prediction is done across time rather than across layers in a hierarchy.

Other methods of training the second-order network involve thinking of it as a Reality Model. For instance, Bayesian methods in deep learning view perception as an inference process, in which a neural network attempts to infer the values of latent variables that might have generated the data. For ideal Bayesian inference, these latent variables must be sampled according to their posterior probability, which is proportional to their prior probability multiplied by their likelihood of having generated the data under some Reality Model that specifies how latents produce sensory observations. Thus, the inference machinery for latent variable values can be seen as producing perceptual first-order representations, while the prior and likelihood can be seen as the outputs of second-order networks that score the probability that the first-order representations are true:

\begin{equation}
P(Latent\:Variables\:|\:Data)\:error\propto P(Latent\:Variables)\times P(Data\:|\:Latent\:Variables)
\nonumber
\end{equation}

Crucially, the division into two systems - one (first-order) doing approximate inference and the other (second-order) providing an unnormalized posterior probability - is because sampling according to the true Bayesian posterior is intractable in general. It can therefore be advantageous to generate candidate samples using an approximate sampling method and then judge the quality of these samples using a Reality Model that consists of a prior and likelihood function. Several approximate Bayesian sampling methods exist. For instance, a recent class of models called Generative Flow Networks (GFlowNets) provides a framework for doing approximate Bayesian inference using modern deep neural networks \cite{Bengio2021-ga, Bengio2021-nb}, and can even be used to jointly train the inference model (first-order network) at the same time as the Reality Model (second-order network) \cite{Hu2023-jx, Zhang2022-ho}.

A final possibility is to learn a Reality Model implicitly through adversarial methods. In particular, Generative Adversarial Networks (GANs) \cite{Goodfellow2014-az} are a class of methods in which a generator network attempts to sample synthetic data and a discriminator attempts to differentiate real data from synthetic data produced by the generator. The generator is optimized to fool the discriminator into predicting its outputs as ‘real’, with the result that, if the networks used are powerful enough, the stable solution is for the generator to sample data from the true data distribution. An implementation might involve (a) a first-order perception network $N_1$ that produces perceptual representations given sensory data $S$ and generates perceptual representation $h(S)=N_1(S)$, (b) a generator network G that inverts this process by first sampling synthetic perceptual representations z and then producing synthetic sensory data given those percepts $h(S)=G(z)$, and (c) a second-order network $D$ that is trained to discriminate between the real sensory-percept tuples produced by the first-order network $h(S)$ and the synthetic ones produced by the generator $h(S)$: $D(x)=P(real|x)$, where $x$ is a synthetic or real input (adapted from \cite{Gershman2019-dc}.

\section{Implementation of  Qualitative Representations}\label{sec6}

\begin{figure}[h]
\centering
\includegraphics[width=0.9\textwidth]{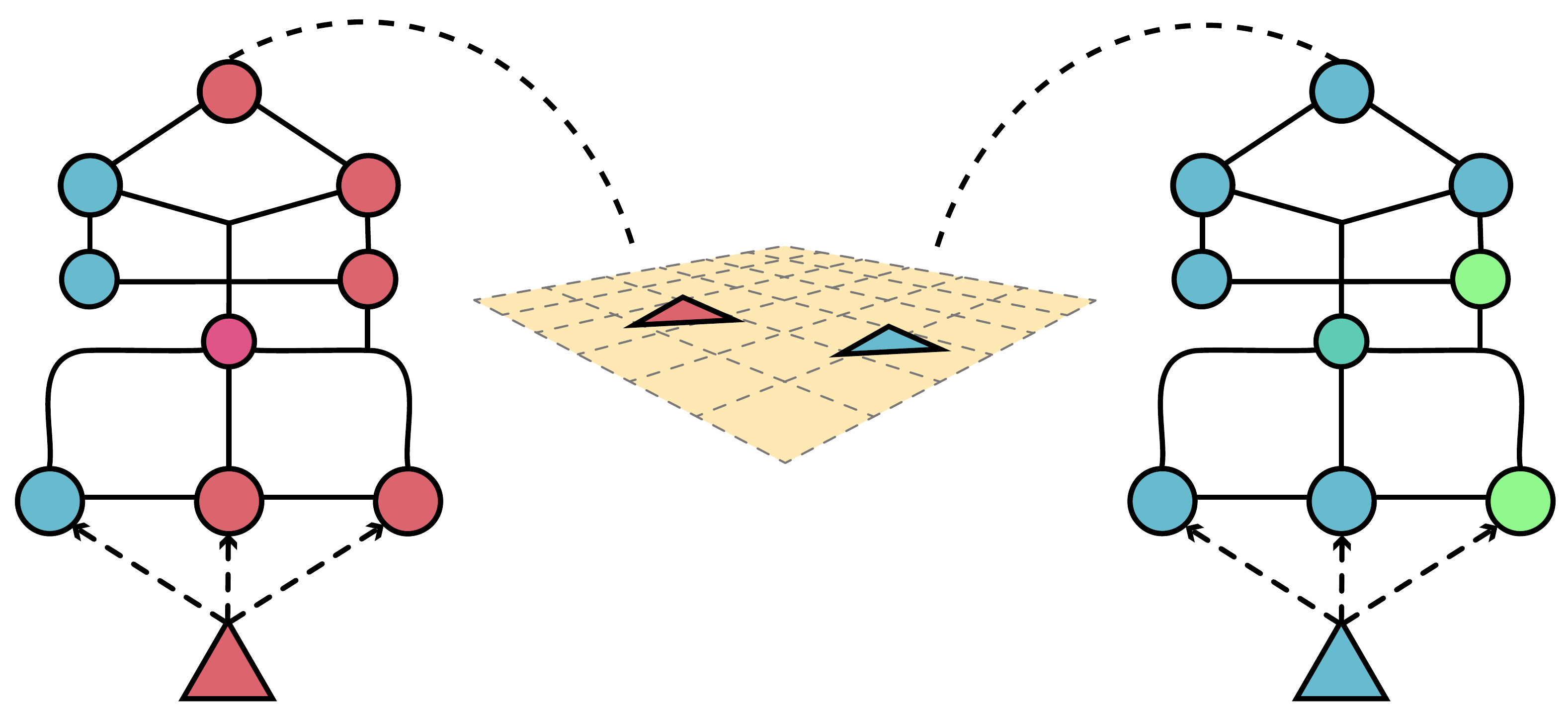}
\caption{Schematic representation of how the degree of similarity between two activity patterns in a neuronal network can be assessed.}\label{fig3}
\end{figure}

For functional sentience to emerge, the Central Processing System needs to have some access to information about fine-grained sensory signal contents. Recall that according to our definition, for sentience to occur, the relevant sensory signals need to be qualitative. One of the most important features of all deep neural networks is that each layer endows the model with a smooth representation space. This means that in neural networks with perceptual functions, the activations already exist in some representation space $X$ \cite{Mamou2020-mw, Allen2019-ub, Bengio2013-ou, Bojanowski2017-dl}, such that the degree of similarity between any two activity patterns can be easily calculated as the Euclidean distance between two points $S_i$ and $S_j$ representing the two stimuli in that space (Fig. \ref{fig3}):

\begin{equation}
d(S_i,S_j)= S_i - S_j
\nonumber
\end{equation}

In a discrete space, the distance might be defined by the shortest path between nodes in a graph or another appropriate metric for the structure. The smoothness of these representation spaces is thought to be one of the primary reasons why they can generalize to novel inputs at test time. Even if a novel input drives a layer into a different activation pattern, this pattern can still be interpreted by subsequent layers so long as it is within the model’s training distribution (that is, so long as the layer activated with similar patterns at training time) \cite{Belkin2018-lh, Bartlett2021-jm}. 

This way, within such a representation space $X$, a stimulus is represented in terms of how similar it is to all other stimuli within that space. By knowing the location of the stimuli in this space, we know its Similarity Profile. In fact, there is substantial evidence that the perceptual representation spaces learned by current deep neural networks already closely resemble those of the human brain. For instance, Kriegeskorte finds that the matrix of pairwise dissimilarities between representations of many visual stimuli in a deep neural network is closely aligned to what is observed in neural activity and that there is even a hierarchical correspondence between successive layers in the network and the hierarchical organization of the visual cortex \cite{Kriegeskorte2015-eq}.

However, just because such spaces could be constructed does not mean that the Central Processing System will have direct access to all fine-grained details of stimuli. In humans, such qualitative information of subjective experience does not necessarily exist somewhere in our brains in some decodable, explicit format \cite{Malach2021-nh, Lau2022-sg}. Rather, we may have implicit access to the information about stimuli similarity. This means it is not computed by the Central Processing System from fine-grained details on demand but is pre-computed and explicitly represented to the Central Processing System in the form of a Similarity Profile.

How can this kind of direct access be implemented in an AI agent? One possibility is to make a signal available to the Central Processing System only in the form of the location of the representation space that captures the Similarity Profile information for the relevant stimulus. To simply formalize it, let $SP(S)$ represent the Similarity Profile for a stimulus $S$ within the representation space $X$. The Central Processing System can then access $SP(S)$ directly by referencing the location $S$ within $X$, where $SP(S)$ is stored as a vector, matrix, or another structured format depending on the space. However, it can’t reference and assess the internal states of $S$ directly.

Therefore, the Central Processing System registers certain early sensory signals only in the form of its location of the representation space that is associated with the actual signal of a proxy. This Similarity Profile information, besides being available to the Central Processing System, also satisfies the other criterion for being subjective, in that it is based on the agent’s own sensory signals and is also automatic in the sense that it doesn’t require an effortful cognitive process. This implies that the Similarity Profile $SP(S)$ is automatically generated as part of the perceptual processing pipeline and is instantly available to the Central Processing System as soon as the sensory data $S$ is processed.

\section{Implications}\label{sec7}

The above architecture is just a sketch; there are certainly other alternatives and possibly even better ones. The purpose here is to illustrate that our approach to sentience is useful enough for designing AI systems. Ultimately, most of us are concerned with machine sentience because we want to understand what it means for an agent to have subjective experiences as we do. These issues are tied to ethical considerations because we want to know \textit{a priori} whether certain agents can be potential victims of subjective suffering before we end up building them.

Let’s consider the example of pain to see if assertoric and qualitative properties of signals are each necessary and jointly sufficient. Imagine that an agent detects a signal that is labeled as a ‘sharp pain’, but has no idea whether this is subjectively more similar to a dull pain, an itch, or a gentle stroke by a feather. An agent doesn’t discriminate between them, meaning that it doesn’t know if the pain is any different from any other sensory signal. In this case, even if the agent recognizes that something is happening, this is just not qualitatively the kind of feeling that we know and care about. In formal terms, the agent lacks a well-defined similarity structure that allows it to place this pain signal relative to other sensory experiences.

Or, the agent may be able to reason through linguistic knowledge that a sharp pain is generally considered to be more similar to a dull pain than to a gentle stroke by other humans. But then there is no sense in which the signal is being processed subjectively in ways akin to how we experience pain, it lacks an internal representation. 

Alternatively, we can consider a scenario where the signal is qualitative but not assertoric. In this case, upon detecting the pain signal, the agent may check whether there is some damage at the bodily location and if it turns out, there is no damage at all, the pain signal would then stop making any impact on the Central Reasoning System. Again, this would just not be the same kind of pain \textit{as we know it}.

On the other hand, if a machine is to actually have ‘pain’ signals that are both assertoric and qualitative, it would be quite hard to imagine that they wouldn’t be very distressing from the point of view of the agent. Formally, this could be represented by ensuring that the pain signal $p$ is both non-ignorable $\alpha(p)\gg\epsilon$ and embedded in a well-defined similarity structure of space $X$, where:
\begin{equation}
p\in{X},\alpha(p)\:is\:such\:that\:d(p,\:any\:other\:sensory\:signal)>\delta
\nonumber
\end{equation}

Here, $alpha(p)$ represents the intensity or priority of the pain signal $p$, $\epsilon$ is a threshold for non-ignorable signals, and $\delta$ represents a threshold for distinctiveness within the similarity structure. While we do not argue here that such a machine would be fully conscious in ways that we are, we submit that it satisfies some useful and meaningful notion of sentience.

The fact that it seems relatively easy to build sentient AI in the functional sentience paradigm may be concerning. However, once we know what it takes for sentience to happen, this can also help us avoid building such agents inadvertently. Sentience is also specific to the types of sensory signals and carries specific functional benefits that we don’t discuss in detail here. While it may be advantageous to allow some agents to have color sentience, we can also specifically design for them to never have sentient pain, for example. The ethical considerations for machines are still in their infancy. We hope that by fleshing out a clear roadmap, experts can navigate accordingly with care, balancing between functional advantages and risks.

\bmhead{Acknowledgements}

The authors express their gratitude to Matthias Michel and Megan Peters for their generous discussion and review of the manuscript.

\bmhead{Conflict of interest}

The authors declare no conflict of interest.


\begin{thebibliography}{44}
\ifx \bisbn   \undefined \def \bisbn  #1{ISBN #1}\fi
\ifx \binits  \undefined \def \binits#1{#1}\fi
\ifx \bauthor  \undefined \def \bauthor#1{#1}\fi
\ifx \batitle  \undefined \def \batitle#1{#1}\fi
\ifx \bjtitle  \undefined \def \bjtitle#1{#1}\fi
\ifx \bvolume  \undefined \def \bvolume#1{\textbf{#1}}\fi
\ifx \byear  \undefined \def \byear#1{#1}\fi
\ifx \bissue  \undefined \def \bissue#1{#1}\fi
\ifx \bfpage  \undefined \def \bfpage#1{#1}\fi
\ifx \blpage  \undefined \def \blpage #1{#1}\fi
\ifx \burl  \undefined \def \burl#1{\textsf{#1}}\fi
\ifx \doiurl  \undefined \def \doiurl#1{\url{https://doi.org/#1}}\fi
\ifx \betal  \undefined \def \betal{\textit{et al.}}\fi
\ifx \binstitute  \undefined \def \binstitute#1{#1}\fi
\ifx \binstitutionaled  \undefined \def \binstitutionaled#1{#1}\fi
\ifx \bctitle  \undefined \def \bctitle#1{#1}\fi
\ifx \beditor  \undefined \def \beditor#1{#1}\fi
\ifx \bpublisher  \undefined \def \bpublisher#1{#1}\fi
\ifx \bbtitle  \undefined \def \bbtitle#1{#1}\fi
\ifx \bedition  \undefined \def \bedition#1{#1}\fi
\ifx \bseriesno  \undefined \def \bseriesno#1{#1}\fi
\ifx \blocation  \undefined \def \blocation#1{#1}\fi
\ifx \bsertitle  \undefined \def \bsertitle#1{#1}\fi
\ifx \bsnm \undefined \def \bsnm#1{#1}\fi
\ifx \bsuffix \undefined \def \bsuffix#1{#1}\fi
\ifx \bparticle \undefined \def \bparticle#1{#1}\fi
\ifx \barticle \undefined \def \barticle#1{#1}\fi
\bibcommenthead
\ifx \bconfdate \undefined \def \bconfdate #1{#1}\fi
\ifx \botherref \undefined \def \botherref #1{#1}\fi
\ifx \url \undefined \def \url#1{\textsf{#1}}\fi
\ifx \bchapter \undefined \def \bchapter#1{#1}\fi
\ifx \bbook \undefined \def \bbook#1{#1}\fi
\ifx \bcomment \undefined \def \bcomment#1{#1}\fi
\ifx \oauthor \undefined \def \oauthor#1{#1}\fi
\ifx \citeauthoryear \undefined \def \citeauthoryear#1{#1}\fi
\ifx \endbibitem  \undefined \def \endbibitem {}\fi
\ifx \bconflocation  \undefined \def \bconflocation#1{#1}\fi
\ifx \arxivurl  \undefined \def \arxivurl#1{\textsf{#1}}\fi
\csname PreBibitemsHook\endcsname

\bibitem[\protect\citeauthoryear{Roelofs}{2021}]{Roelofs2021-qm}
\begin{botherref}
\oauthor{\bsnm{Roelofs}, \binits{L.}}:
Is panpsychism at odds with science?
J. Conscious. Stud.
(2021)
\end{botherref}
\endbibitem

\bibitem[\protect\citeauthoryear{Fleming et~al.}{2023}]{Fleming2023-nf}
\begin{botherref}
\oauthor{\bsnm{Fleming}, \binits{S.}},
\oauthor{\bsnm{Frith}, \binits{C.D.}},
\oauthor{\bsnm{Goodale}, \binits{M.}},
\oauthor{\bsnm{Lau}, \binits{H.}},
\oauthor{\bsnm{LeDoux}, \binits{J.E.}},
\oauthor{\bsnm{Lee}, \binits{A.L.F.}},
\oauthor{\bsnm{Michel}, \binits{M.}},
\oauthor{\bsnm{Owen}, \binits{A.M.}},
\oauthor{\bsnm{Peters}, \binits{M.A.K.}},
\oauthor{\bsnm{Slagter}, \binits{H.A.}},
\oauthor{\bsnm{{Others}}}:
The integrated information theory of consciousness as pseudoscience
(2023)
\end{botherref}
\endbibitem

\bibitem[\protect\citeauthoryear{Dehaene et~al.}{2011}]{Dehaene2011-jr}
\begin{bchapter}
\bauthor{\bsnm{Dehaene}, \binits{S.}},
\bauthor{\bsnm{Changeux}, \binits{J.-P.}},
\bauthor{\bsnm{Naccache}, \binits{L.}}:
\bctitle{The global neuronal workspace model of conscious access: From neuronal architectures to clinical applications}.
In: \bbtitle{Research and Perspectives in Neurosciences}.
\bsertitle{Research and perspectives in neurosciences},
pp. \bfpage{55}--\blpage{84}.
\bpublisher{Springer},
\blocation{Berlin, Heidelberg}
(\byear{2011})
\end{bchapter}
\endbibitem

\bibitem[\protect\citeauthoryear{Butlin et~al.}{2023}]{Butlin2023-jf}
\begin{botherref}
\oauthor{\bsnm{Butlin}, \binits{P.}},
\oauthor{\bsnm{Long}, \binits{R.}},
\oauthor{\bsnm{Elmoznino}, \binits{E.}},
\oauthor{\bsnm{Bengio}, \binits{Y.}},
\oauthor{\bsnm{Birch}, \binits{J.}},
\oauthor{\bsnm{Constant}, \binits{A.}},
\oauthor{\bsnm{Deane}, \binits{G.}},
\oauthor{\bsnm{Fleming}, \binits{S.M.}},
\oauthor{\bsnm{Frith}, \binits{C.}},
\oauthor{\bsnm{Ji}, \binits{X.}},
\oauthor{\bsnm{Kanai}, \binits{R.}},
\oauthor{\bsnm{Klein}, \binits{C.}},
\oauthor{\bsnm{Lindsay}, \binits{G.}},
\oauthor{\bsnm{Michel}, \binits{M.}},
\oauthor{\bsnm{Mudrik}, \binits{L.}},
\oauthor{\bsnm{Peters}, \binits{M.A.K.}},
\oauthor{\bsnm{Schwitzgebel}, \binits{E.}},
\oauthor{\bsnm{Simon}, \binits{J.}},
\oauthor{\bsnm{VanRullen}, \binits{R.}}:
Consciousness in artificial intelligence: Insights from the science of consciousness.
arXiv [cs.AI]
(2023)
\end{botherref}
\endbibitem

\bibitem[\protect\citeauthoryear{Baars}{1993}]{Baars1993-lk}
\begin{bbook}
\bauthor{\bsnm{Baars}, \binits{B.J.}}:
\bbtitle{A Cognitive Theory of Consciousness}.
\bpublisher{Cambridge University Press},
\blocation{Cambridge, England}
(\byear{1993})
\end{bbook}
\endbibitem

\bibitem[\protect\citeauthoryear{Kay et~al.}{2024}]{Kay2024-rb}
\begin{barticle}
\bauthor{\bsnm{Kay}, \binits{L.}},
\bauthor{\bsnm{Keogh}, \binits{R.}},
\bauthor{\bsnm{Pearson}, \binits{J.}}:
\batitle{Slower but more accurate mental rotation performance in aphantasia linked to differences in cognitive strategies}.
\bjtitle{Conscious. Cogn.}
\bvolume{121}(\bissue{103694}),
\bfpage{103694}
(\byear{2024})
\end{barticle}
\endbibitem

\bibitem[\protect\citeauthoryear{Zhao et~al.}{2022}]{Zhao2022-fe}
\begin{barticle}
\bauthor{\bsnm{Zhao}, \binits{B.}},
\bauthor{\bsnm{Della~Sala}, \binits{S.}},
\bauthor{\bsnm{Zeman}, \binits{A.}},
\bauthor{\bsnm{Gherri}, \binits{E.}}:
\batitle{Spatial transformation in mental rotation tasks in aphantasia}.
\bjtitle{Psychon. Bull. Rev.}
\bvolume{29}(\bissue{6}),
\bfpage{2096}--\blpage{2107}
(\byear{2022})
\end{barticle}
\endbibitem

\bibitem[\protect\citeauthoryear{Pounder et~al.}{2022}]{Pounder2022-uc}
\begin{barticle}
\bauthor{\bsnm{Pounder}, \binits{Z.}},
\bauthor{\bsnm{Jacob}, \binits{J.}},
\bauthor{\bsnm{Evans}, \binits{S.}},
\bauthor{\bsnm{Loveday}, \binits{C.}},
\bauthor{\bsnm{Eardley}, \binits{A.F.}},
\bauthor{\bsnm{Silvanto}, \binits{J.}}:
\batitle{Only minimal differences between individuals with congenital aphantasia and those with typical imagery on neuropsychological tasks that involve imagery}.
\bjtitle{Cortex}
\bvolume{148},
\bfpage{180}--\blpage{192}
(\byear{2022})
\end{barticle}
\endbibitem

\bibitem[\protect\citeauthoryear{Michel et~al.}{2025}]{Michel2025-cj}
\begin{barticle}
\bauthor{\bsnm{Michel}, \binits{M.}},
\bauthor{\bsnm{Morales}, \binits{J.}},
\bauthor{\bsnm{Block}, \binits{N.}},
\bauthor{\bsnm{Lau}, \binits{H.}}:
\batitle{Aphantasia as imagery blindsight}.
\bjtitle{Trends Cogn. Sci.}
\bvolume{29}(\bissue{1}),
\bfpage{8}--\blpage{9}
(\byear{2025})
\end{barticle}
\endbibitem

\bibitem[\protect\citeauthoryear{Weiskrantz}{1997}]{Weiskrantz1997-kx}
\begin{bbook}
\bauthor{\bsnm{Weiskrantz}, \binits{L.}}:
\bbtitle{Consciousness Lost and Found: A Neuropsychological Exploration}.
\bpublisher{Oxford University Press},
\blocation{Cary, NC}
(\byear{1997})
\end{bbook}
\endbibitem

\bibitem[\protect\citeauthoryear{Aleci and Dutto}{2024}]{Aleci2024-mh}
\begin{botherref}
\oauthor{\bsnm{Aleci}, \binits{C.}},
\oauthor{\bsnm{Dutto}, \binits{K.}}:
Seeing the invisible: theory and evidence of blindsight.
Discov Med
\textbf{1}(1)
(2024)
\end{botherref}
\endbibitem

\bibitem[\protect\citeauthoryear{Ajina and Bridge}{2016}]{Ajina2016-zk}
\begin{barticle}
\bauthor{\bsnm{Ajina}, \binits{S.}},
\bauthor{\bsnm{Bridge}, \binits{H.}}:
\batitle{Blindsight and unconscious vision: What they teach us about the human visual system}.
\bjtitle{Neuroscientist}
\bvolume{23}(\bissue{5}),
\bfpage{529}--\blpage{541}
(\byear{2016})
\end{barticle}
\endbibitem

\bibitem[\protect\citeauthoryear{Overgaard}{2012}]{Overgaard2012-xb}
\begin{barticle}
\bauthor{\bsnm{Overgaard}, \binits{M.}}:
\batitle{Blindsight: recent and historical controversies on the blindness of blindsight}.
\bjtitle{Wiley Interdiscip. Rev. Cogn. Sci.}
\bvolume{3}(\bissue{6}),
\bfpage{607}--\blpage{614}
(\byear{2012})
\end{barticle}
\endbibitem

\bibitem[\protect\citeauthoryear{Polger}{2012}]{Polger2012-sl}
\begin{barticle}
\bauthor{\bsnm{Polger}, \binits{T.W.}}:
\batitle{Functionalism as a philosophical theory of the cognitive sciences: Functionalism as a philosophical theory}.
\bjtitle{Wiley Interdiscip. Rev. Cogn. Sci.}
\bvolume{3}(\bissue{3}),
\bfpage{337}--\blpage{348}
(\byear{2012})
\end{barticle}
\endbibitem

\bibitem[\protect\citeauthoryear{Papamarkou et~al.}{2024}]{Papamarkou2024-ig}
\begin{botherref}
\oauthor{\bsnm{Papamarkou}, \binits{T.}},
\oauthor{\bsnm{Skoularidou}, \binits{M.}},
\oauthor{\bsnm{Palla}, \binits{K.}},
\oauthor{\bsnm{Aitchison}, \binits{L.}},
\oauthor{\bsnm{Arbel}, \binits{J.}},
\oauthor{\bsnm{Dunson}, \binits{D.}},
\oauthor{\bsnm{Filippone}, \binits{M.}},
\oauthor{\bsnm{Fortuin}, \binits{V.}},
\oauthor{\bsnm{Hennig}, \binits{P.}},
\oauthor{\bsnm{Hernández-Lobato}, \binits{J.M.}},
\oauthor{\bsnm{Hubin}, \binits{A.}},
\oauthor{\bsnm{Immer}, \binits{A.}},
\oauthor{\bsnm{Karaletsos}, \binits{T.}},
\oauthor{\bsnm{Khan}, \binits{M.E.}},
\oauthor{\bsnm{Kristiadi}, \binits{A.}},
\oauthor{\bsnm{Li}, \binits{Y.}},
\oauthor{\bsnm{Mandt}, \binits{S.}},
\oauthor{\bsnm{Nemeth}, \binits{C.}},
\oauthor{\bsnm{Osborne}, \binits{M.A.}},
\oauthor{\bsnm{Rudner}, \binits{T.G.J.}},
\oauthor{\bsnm{Rügamer}, \binits{D.}},
\oauthor{\bsnm{Teh}, \binits{Y.W.}},
\oauthor{\bsnm{Welling}, \binits{M.}},
\oauthor{\bsnm{Wilson}, \binits{A.G.}},
\oauthor{\bsnm{Zhang}, \binits{R.}}:
Position: Bayesian deep learning is needed in the age of large-scale {AI}.
arXiv [cs.LG]
(2024)
\end{botherref}
\endbibitem

\bibitem[\protect\citeauthoryear{Dijkstra et~al.}{2022}]{Dijkstra2022-ty}
\begin{botherref}
\oauthor{\bsnm{Dijkstra}, \binits{N.}},
\oauthor{\bsnm{Kok}, \binits{P.}},
\oauthor{\bsnm{Fleming}, \binits{S.M.}}:
Perceptual reality monitoring: Neural mechanisms dissociating imagination from reality.
Neurosci. Biobehav. Rev.
\textbf{135}(104557)
(2022)
\end{botherref}
\endbibitem

\bibitem[\protect\citeauthoryear{Muller-Lyer}{1889}]{Muller-Lyer1889-qt}
\begin{barticle}
\bauthor{\bsnm{Muller-Lyer}, \binits{F.C.}}:
\batitle{Optische urteilstauschungen}.
\bjtitle{Archiv fur Anatomie und Physiologie, Physiologische Abteilung}
\bvolume{2},
\bfpage{263}--\blpage{270}
(\byear{1889})
\end{barticle}
\endbibitem

\bibitem[\protect\citeauthoryear{Beck}{2019}]{Beck2019-jn}
\begin{botherref}
\oauthor{\bsnm{Beck}, \binits{J.}}:
Perception is analog: The argument from weber's law.
J. Philos.
(2019)
\end{botherref}
\endbibitem

\bibitem[\protect\citeauthoryear{Clark}{2000}]{Clark2000-ki}
\begin{bbook}
\bauthor{\bsnm{Clark}, \binits{A.}}:
\bbtitle{A Theory of Sentience}.
\bpublisher{Clarendon Press},
\blocation{Oxford, England}
(\byear{2000})
\end{bbook}
\endbibitem

\bibitem[\protect\citeauthoryear{Marjieh et~al.}{2023}]{Marjieh2023-jz}
\begin{botherref}
\oauthor{\bsnm{Marjieh}, \binits{R.}},
\oauthor{\bsnm{Sucholutsky}, \binits{I.}},
\oauthor{\bsnm{Rijn}, \binits{P.}},
\oauthor{\bsnm{Jacoby}, \binits{N.}},
\oauthor{\bsnm{Griffiths}, \binits{T.}}:
Large language models predict human sensory judgments across six modalities.
Sci. Rep.
\textbf{14}
(2023)
\end{botherref}
\endbibitem

\bibitem[\protect\citeauthoryear{Kawakita et~al.}{2024}]{Kawakita2024-cp}
\begin{barticle}
\bauthor{\bsnm{Kawakita}, \binits{G.}},
\bauthor{\bsnm{Zeleznikow-Johnston}, \binits{A.}},
\bauthor{\bsnm{Tsuchiya}, \binits{N.}},
\bauthor{\bsnm{Oizumi}, \binits{M.}}:
\batitle{Gromov-wasserstein unsupervised alignment reveals structural correspondences between the color similarity structures of humans and large language models}.
\bjtitle{Sci. Rep.}
\bvolume{14}(\bissue{1}),
\bfpage{15917}
(\byear{2024})
\end{barticle}
\endbibitem

\bibitem[\protect\citeauthoryear{Feng et~al.}{2024}]{Feng2024-vz}
\begin{botherref}
\oauthor{\bsnm{Feng}, \binits{T.}},
\oauthor{\bsnm{Jin}, \binits{C.}},
\oauthor{\bsnm{Liu}, \binits{J.}},
\oauthor{\bsnm{Zhu}, \binits{K.}},
\oauthor{\bsnm{Tu}, \binits{H.}},
\oauthor{\bsnm{Cheng}, \binits{Z.}},
\oauthor{\bsnm{Lin}, \binits{G.}},
\oauthor{\bsnm{You}, \binits{J.}}:
How far are we from {AGI}: Are {LLMs} all we need?
arXiv [cs.AI]
(2024)
\end{botherref}
\endbibitem

\bibitem[\protect\citeauthoryear{Morris et~al.}{2023}]{Morris2023-xf}
\begin{botherref}
\oauthor{\bsnm{Morris}, \binits{M.R.}},
\oauthor{\bsnm{Sohl-dickstein}, \binits{J.}},
\oauthor{\bsnm{Fiedel}, \binits{N.}},
\oauthor{\bsnm{Warkentin}, \binits{T.}},
\oauthor{\bsnm{Dafoe}, \binits{A.}},
\oauthor{\bsnm{Faust}, \binits{A.}},
\oauthor{\bsnm{Farabet}, \binits{C.}},
\oauthor{\bsnm{Legg}, \binits{S.}}:
Levels of {AGI} for operationalizing progress on the path to {AGI}.
arXiv [cs.AI]
(2023)
\end{botherref}
\endbibitem

\bibitem[\protect\citeauthoryear{Prabhakar et~al.}{2024}]{Prabhakar2024-eq}
\begin{botherref}
\oauthor{\bsnm{Prabhakar}, \binits{A.}},
\oauthor{\bsnm{Griffiths}, \binits{T.L.}},
\oauthor{\bsnm{McCoy}, \binits{R.T.}}:
Deciphering the factors influencing the efficacy of chain-of-thought: Probability, memorization, and noisy reasoning.
arXiv [cs.CL]
(2024)
\end{botherref}
\endbibitem

\bibitem[\protect\citeauthoryear{Alamia et~al.}{2023}]{Alamia2023-bw}
\begin{barticle}
\bauthor{\bsnm{Alamia}, \binits{A.}},
\bauthor{\bsnm{Mozafari}, \binits{M.}},
\bauthor{\bsnm{Choksi}, \binits{B.}},
\bauthor{\bsnm{VanRullen}, \binits{R.}}:
\batitle{On the role of feedback in image recognition under noise and adversarial attacks: A predictive coding perspective}.
\bjtitle{Neural Netw.}
\bvolume{157},
\bfpage{280}--\blpage{287}
(\byear{2023})
\end{barticle}
\endbibitem

\bibitem[\protect\citeauthoryear{Wei et~al.}{2022}]{Wei2022-cx}
\begin{botherref}
\oauthor{\bsnm{Wei}, \binits{J.}},
\oauthor{\bsnm{Wang}, \binits{X.}},
\oauthor{\bsnm{Schuurmans}, \binits{D.}},
\oauthor{\bsnm{Bosma}, \binits{M.}},
\oauthor{\bsnm{Chi}, \binits{E.H.}},
\oauthor{\bsnm{Xia}, \binits{F.}},
\oauthor{\bsnm{Le}, \binits{Q.}},
\oauthor{\bsnm{Zhou}, \binits{D.}}:
Chain of thought prompting elicits reasoning in large language models.
Neural Inf Process Syst,
24824--24837
(2022)
\end{botherref}
\endbibitem

\bibitem[\protect\citeauthoryear{Chia et~al.}{2023}]{Chia2023-pa}
\begin{botherref}
\oauthor{\bsnm{Chia}, \binits{Y.K.}},
\oauthor{\bsnm{Chen}, \binits{G.}},
\oauthor{\bsnm{Tuan}, \binits{L.A.}},
\oauthor{\bsnm{Poria}, \binits{S.}},
\oauthor{\bsnm{Bing}, \binits{L.}}:
Contrastive chain-of-thought prompting.
arXiv [cs.CL]
(2023)
\end{botherref}
\endbibitem

\bibitem[\protect\citeauthoryear{Millidge et~al.}{2022}]{Millidge2022-to}
\begin{botherref}
\oauthor{\bsnm{Millidge}, \binits{B.}},
\oauthor{\bsnm{Tschantz}, \binits{A.}},
\oauthor{\bsnm{Buckley}, \binits{C.L.}}:
Predictive coding approximates backprop along arbitrary computation graphs.
Neural Comput.
(2022)
\end{botherref}
\endbibitem

\bibitem[\protect\citeauthoryear{van~den Oord et~al.}{2018}]{van-den-Oord2018-wg}
\begin{botherref}
\oauthor{\bsnm{Oord}, \binits{A.}},
\oauthor{\bsnm{Li}, \binits{Y.}},
\oauthor{\bsnm{Vinyals}, \binits{O.}}:
Representation learning with contrastive predictive coding.
arXiv [cs.LG]
(2018)
\end{botherref}
\endbibitem

\bibitem[\protect\citeauthoryear{Bengio et~al.}{2021a}]{Bengio2021-ga}
\begin{botherref}
\oauthor{\bsnm{Bengio}, \binits{Y.}},
\oauthor{\bsnm{Lahlou}, \binits{S.}},
\oauthor{\bsnm{Deleu}, \binits{T.}},
\oauthor{\bsnm{Hu}, \binits{E.J.}},
\oauthor{\bsnm{Tiwari}, \binits{M.}},
\oauthor{\bsnm{Bengio}, \binits{E.}}:
{GFlowNet} foundations.
arXiv [cs.LG]
(2021)
\end{botherref}
\endbibitem

\bibitem[\protect\citeauthoryear{Bengio et~al.}{2021b}]{Bengio2021-nb}
\begin{botherref}
\oauthor{\bsnm{Bengio}, \binits{E.}},
\oauthor{\bsnm{Jain}, \binits{M.}},
\oauthor{\bsnm{Korablyov}, \binits{M.}},
\oauthor{\bsnm{Precup}, \binits{D.}},
\oauthor{\bsnm{Bengio}, \binits{Y.}}:
Flow network based generative models for non-iterative diverse candidate generation.
arXiv [cs.LG]
(2021)
\end{botherref}
\endbibitem

\bibitem[\protect\citeauthoryear{Hu et~al.}{2023}]{Hu2023-jx}
\begin{botherref}
\oauthor{\bsnm{Hu}, \binits{E.J.}},
\oauthor{\bsnm{Malkin}, \binits{N.}},
\oauthor{\bsnm{Jain}, \binits{M.}},
\oauthor{\bsnm{Everett}, \binits{K.}},
\oauthor{\bsnm{Graikos}, \binits{A.}},
\oauthor{\bsnm{Bengio}, \binits{Y.}}:
{GFlowNet}-{EM} for learning compositional latent variable models.
arXiv [cs.LG]
(2023)
\end{botherref}
\endbibitem

\bibitem[\protect\citeauthoryear{Zhang et~al.}{2022}]{Zhang2022-ho}
\begin{botherref}
\oauthor{\bsnm{Zhang}, \binits{D.}},
\oauthor{\bsnm{Chen}, \binits{R.T.Q.}},
\oauthor{\bsnm{Malkin}, \binits{N.}},
\oauthor{\bsnm{Bengio}, \binits{Y.}}:
Unifying generative models with {GFlowNets} and beyond.
arXiv [cs.LG]
(2022)
\end{botherref}
\endbibitem

\bibitem[\protect\citeauthoryear{Goodfellow et~al.}{2014}]{Goodfellow2014-az}
\begin{botherref}
\oauthor{\bsnm{Goodfellow}, \binits{I.}},
\oauthor{\bsnm{Pouget-Abadie}, \binits{J.}},
\oauthor{\bsnm{Mirza}, \binits{M.}},
\oauthor{\bsnm{Xu}, \binits{B.}},
\oauthor{\bsnm{Warde-Farley}, \binits{D.}},
\oauthor{\bsnm{Ozair}, \binits{S.}},
\oauthor{\bsnm{Courville}, \binits{A.}},
\oauthor{\bsnm{Bengio}, \binits{Y.}}:
Generative adversarial nets.
Adv. Neural Inf. Process. Syst.
\textbf{27}
(2014)
\end{botherref}
\endbibitem

\bibitem[\protect\citeauthoryear{Gershman}{2019}]{Gershman2019-dc}
\begin{barticle}
\bauthor{\bsnm{Gershman}, \binits{S.J.}}:
\batitle{The generative adversarial brain}.
\bjtitle{Front. Artif. Intell.}
\bvolume{2},
\bfpage{18}
(\byear{2019})
\end{barticle}
\endbibitem

\bibitem[\protect\citeauthoryear{Mamou et~al.}{2020}]{Mamou2020-mw}
\begin{botherref}
\oauthor{\bsnm{Mamou}, \binits{J.}},
\oauthor{\bsnm{Le}, \binits{H.}},
\oauthor{\bsnm{Del~Rio}, \binits{M.}},
\oauthor{\bsnm{Stephenson}, \binits{C.}},
\oauthor{\bsnm{Tang}, \binits{H.}},
\oauthor{\bsnm{Kim}, \binits{Y.}},
\oauthor{\bsnm{Chung}, \binits{S.}}:
Emergence of separable manifolds in deep language representations.
arXiv [cs.CL]
(2020)
\end{botherref}
\endbibitem

\bibitem[\protect\citeauthoryear{Allen and Hospedales}{2019}]{Allen2019-ub}
\begin{barticle}
\bauthor{\bsnm{Allen}, \binits{C.}},
\bauthor{\bsnm{Hospedales}, \binits{T.M.}}:
\batitle{Analogies explained: Towards understanding word embeddings}.
\bjtitle{ICML}
\bvolume{97},
\bfpage{223}--\blpage{231}
(\byear{2019})
\end{barticle}
\endbibitem

\bibitem[\protect\citeauthoryear{Bengio et~al.}{2013}]{Bengio2013-ou}
\begin{barticle}
\bauthor{\bsnm{Bengio}, \binits{Y.}},
\bauthor{\bsnm{Courville}, \binits{A.}},
\bauthor{\bsnm{Vincent}, \binits{P.}}:
\batitle{Representation learning: A review and new perspectives}.
\bjtitle{IEEE transactions on pattern analysis and machine intelligence}
\bvolume{35}(\bissue{8}),
\bfpage{1798}--\blpage{1828}
(\byear{2013})
\end{barticle}
\endbibitem

\bibitem[\protect\citeauthoryear{Bojanowski et~al.}{2017}]{Bojanowski2017-dl}
\begin{botherref}
\oauthor{\bsnm{Bojanowski}, \binits{P.}},
\oauthor{\bsnm{Joulin}, \binits{A.}},
\oauthor{\bsnm{Lopez-Paz}, \binits{D.}},
\oauthor{\bsnm{Szlam}, \binits{A.}}:
Optimizing the latent space of generative networks.
arXiv [stat.ML]
(2017)
\end{botherref}
\endbibitem

\bibitem[\protect\citeauthoryear{Belkin et~al.}{2018}]{Belkin2018-lh}
\begin{bchapter}
\bauthor{\bsnm{Belkin}, \binits{M.}},
\bauthor{\bsnm{Ma}, \binits{S.}},
\bauthor{\bsnm{Mandal}, \binits{S.}}:
\bctitle{To understand deep learning we need to understand kernel learning}.
In: \bbtitle{International Conference on Machine Learning},
vol. \bseriesno{80},
pp. \bfpage{541}--\blpage{549}.
\bpublisher{PMLR},
\blocation{Stockholm, Sweden}
(\byear{2018})
\end{bchapter}
\endbibitem

\bibitem[\protect\citeauthoryear{Bartlett et~al.}{2021}]{Bartlett2021-jm}
\begin{barticle}
\bauthor{\bsnm{Bartlett}, \binits{P.L.}},
\bauthor{\bsnm{Montanari}, \binits{A.}},
\bauthor{\bsnm{Rakhlin}, \binits{A.}}:
\batitle{Deep learning: a statistical viewpoint}.
\bjtitle{Acta Numer.}
\bvolume{30},
\bfpage{87}--\blpage{201}
(\byear{2021})
\end{barticle}
\endbibitem

\bibitem[\protect\citeauthoryear{Kriegeskorte}{2015}]{Kriegeskorte2015-eq}
\begin{barticle}
\bauthor{\bsnm{Kriegeskorte}, \binits{N.}}:
\batitle{Deep neural networks: A new framework for modeling biological vision and brain information processing}.
\bjtitle{Annu. Rev. Vis. Sci.}
\bvolume{1},
\bfpage{417}--\blpage{446}
(\byear{2015})
\end{barticle}
\endbibitem

\bibitem[\protect\citeauthoryear{Malach}{2021}]{Malach2021-nh}
\begin{botherref}
\oauthor{\bsnm{Malach}, \binits{R.}}:
Local neuronal relational structures underlying the contents of human conscious experience.
Neurosci. Conscious.
(2021)
\end{botherref}
\endbibitem

\bibitem[\protect\citeauthoryear{Lau et~al.}{2022}]{Lau2022-sg}
\begin{barticle}
\bauthor{\bsnm{Lau}, \binits{H.}},
\bauthor{\bsnm{Michel}, \binits{M.}},
\bauthor{\bsnm{LeDoux}, \binits{J.E.}},
\bauthor{\bsnm{Fleming}, \binits{S.}}:
\batitle{The mnemonic basis of subjective experience}.
\bjtitle{Nat. Rev. Psychol.}
\bvolume{1},
\bfpage{479}--\blpage{488}
(\byear{2022})
\end{barticle}
\endbibitem

\end{thebibliography}


\end{document}